\documentclass[runningheads]{llncs}
\usepackage[T1]{fontenc}
\usepackage{graphicx}
\usepackage{todonotes}
\usepackage{listings}
\usepackage{multirow}
\usepackage{tabularx}
\usepackage{array}
\usepackage{booktabs}
\usepackage{amsmath}
\usepackage{amssymb}
\usepackage{appendix}
\usepackage{longtable}
\usepackage{comment}
\usepackage{hyperref}
\usepackage{cleveref}
\usepackage[linesnumbered,ruled,vlined]{algorithm2e}
\lstdefinestyle{jsonstyle}{
    language=JSON,
    basicstyle=\ttfamily\small,
    numbers=none,
    backgroundcolor=\color{lightgray},
    showstringspaces=false,
    breaklines=true,
    frame=single,
    captionpos=b,
    commentstyle=\color{gray},
    keywordstyle=\color{blue}
}

\begin{document}
\title{Benchmarking Knowledge Editing using Logical Rules}

\author{
Tatiana Moteu Ngoli \and 
N'Dah Jean Kouagou \and 
Hamada M. Zahera \and 
Axel-Cyrille Ngonga Ngomo
}

\institute{
Data Science Group, Heinz Nixdorf Institute, Paderborn University\\
\email{
tatiana.moteu@upb.de, 
jean.kouagou@upb.de, 
hamada.zahera@upb.de, 
axel.ngonga@upb.de
}
}

\authorrunning{Tatiana Moteu Ngoli et al.}

\maketitle              

\begin{abstract}
Large Language Models (LLMs) are increasingly deployed in real-world applications that require access to up-to-date knowledge. However, retraining LLMs is computationally expensive. Therefore, knowledge editing techniques are crucial for maintaining current information and correcting erroneous assertions within pre-trained models. Current benchmarks for knowledge editing primarily focus on recalling edited facts, often neglecting their logical consequences.
To address this limitation, we introduce a new benchmark designed to evaluate how knowledge editing methods handle the logical consequences of a single fact edit. Our benchmark extracts relevant logical rules from a knowledge graph for a given edit. Then, it generates multi-hop questions based on these rules to assess the impact on logical consequences. Our findings indicate that while existing knowledge editing approaches can accurately insert direct assertions into LLMs, they frequently fail to inject entailed knowledge. Specifically, experiments with popular methods like ROME and FT reveal a substantial performance gap, up to 24\%, between evaluations on directly edited knowledge and on entailed knowledge. This highlights the critical need for semantics-aware evaluation frameworks in knowledge editing.
\end{abstract}


\section{Introduction }
Knowledge editing (KE) in large language models (LLMs) aims to update outdated or incorrect information without retraining the entire model. 
This task is critical because current LLMs cannot automatically adapt to new information, and retraining them is computational expensive~\cite{wei2022chain,valmeekam2022large}. 
Existing studies have primarily focused on two types of methods for knowledge editing: \textit{(i) parameter-based methods} (e.g., ROME~\cite{meng2022romeb}, MEMIT~\cite{meng2022memit}), which locate the relevant parameters associated with specific knowledge within LLMs and update them to incorporate new information, and \textit{(ii) memory-based methods} (e.g., MeLLo~\cite{zhong2024mquakeassessingknowledgeediting}, PokeMQA~\cite{gu2024pokemqa}), which use an external memory to store new knowledge for editing without modifying the pre-trained weights, thereby preserving existing knowledge in the LLM. 
While current benchmarks for KE approaches primarily assess direct edits~\cite{yao2023editinglargelanguagemodels}, a significant challenge remains in evaluating how editing one piece of knowledge affects related information. This is often referred to as correlated or multi-hop knowledge \cite{wu2023evakellmnewbenchmarkevaluating,wei-etal-2025-mlake}. For example, consider updating a language model with the information that Donald Trump is the current President of the United States. While this direct edit ensures the model correctly answers \textit{"Who is the President of the United States?"} with \textit{"Donald Trump,"} it is equally crucial that the model's responses to related queries reflect this edit. For instance, questions like \textit{"Who is the spouse of the U.S. President?"} should now yield \textit{"Melania Trump,"} and \textit{"Who is the Vice President of the United States?"} should provide the appropriate answer corresponding to Trump's administration. These examples highlight the necessity of evaluating whether edits propagate coherently across interconnected facts within the model. 
Different studies have shown that current KE methods can accurately recall edited assertions~\cite{zhang2025locatetheneditmultihopfactualrecall}. However, they often fail to maintain consistency across related information~\cite{rae2022scalinglanguagemodelsmethods,creswell2022selectioninferenceexploitinglargelanguage}, highlighting the need for more comprehensive evaluation frameworks. 
The MQuAKE~\cite{zhong2024mquakeassessingknowledgeediting} benchmark addresses this limitation by assessing models on multi-hop questions that require reasoning over chains of facts affected by a single edit. Yet, MQuAKE relies on manually curated multi-hop questions, which limits its scalability and coverage for cross-domain knowledge edits.
\begin{figure*}[t!]
 \centering
     \includegraphics[width=0.92\textwidth, trim=5 10 5 5, clip]{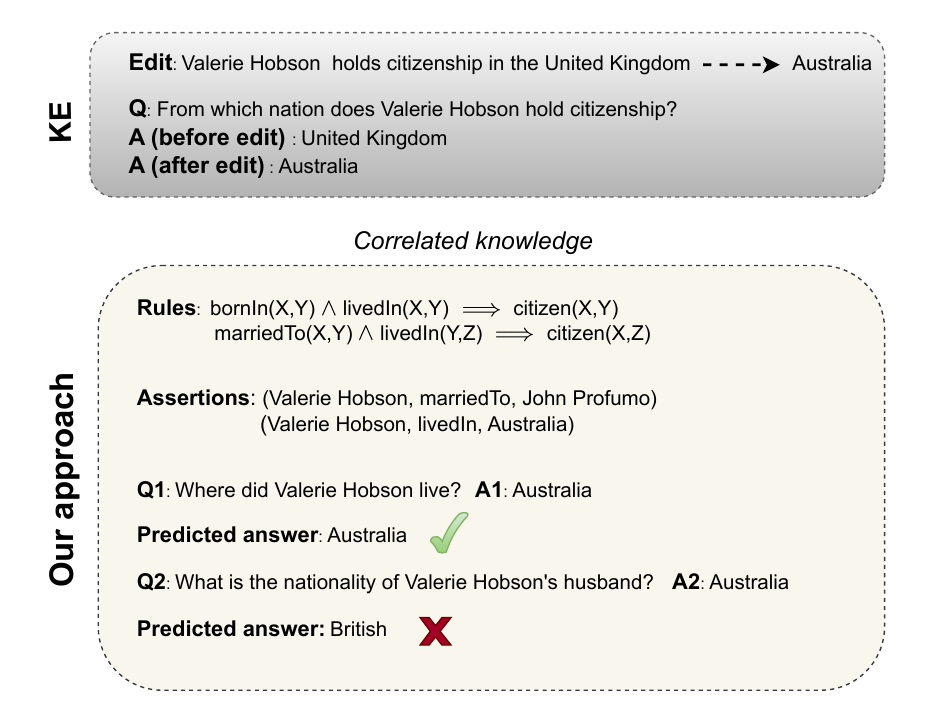}
     \caption{An Example of Single Edit with Correlated Knowledge.}
     \label{fig:approach}
 \end{figure*}
 
In this paper, we propose a novel benchmark that evaluates not only the retention of directly edited assertions, but also the model’s ability to infer and reason over knowledge that is logically entailed by those edits. 
Unlike domain-specific benchmarks, which require costly manual curation and cannot be easily ported across domains or datasets, our method automatically generate test cases for any domain based on logical rules, ensuring scalability and adaptability. In particular, our approach uses AMIE3~\cite{10.1145/2488388.2488425}--a rule-mining system for knowledge graphs--to automatically extract logical relationships and generate test cases. These rules guide the LLM (e.g., GPT-4) to generate multi-hop questions as test cases that verify whether edits to a base assertion propagate to correlated knowledge. Figure~\ref{fig:approach} illustrates this process: editing an assertion (e.g., \textit{updating Valerie Hobson's citizenship from the United Kingdom} to \textit{Australia}) requires corresponding updates to correlated assertions (e.g., \textit{the nationality of Valerie Hobson’s husband}) through logical rules. When a direct assertion is modified, our benchmark evaluates whether the KE method propagates this change to dependent knowledge by generating multi-hop questions (e.g., \textit{What is the nationality of Valerie Hobson’s husband?}) using mined logical relationships (e.g., \textit{spouses typically share citizenship}). 
Our evaluation results show a significant difference in KE effectiveness between direct and correlated knowledge evaluations, particularly in reasoning with correlated knowledge. By automating test cases generation, we enable robust, low-cost evaluation of how KE methods handle reasoning chains, exposing weaknesses in real-world scenarios where assertions are interdependent. 
The main contributions of this paper are summarized as follows:
\begin{itemize}
    \item We propose a new benchmark approach for KE that leverages logical rules to augment correlated knowledge in benchmarking KE. We also perform an evaluation of existing KE methods using two existing datasets.
    \item 
    We conduct several experiments to evaluate different KE methods on test cases generated based on logical rules. 
    
    \item 
    We release our benchmark framework as open-source on the GitHub repository\footnote{\url{https://github.com/dice-group/Benchmarking-KE}} for reproducing our experiments and supporting further research in knowledge editing.  
\end{itemize}

\section{Related Work}
%In this section, we present an overview of knowledge editing approaches and their existing benchmarks.

\subsection{An Overview of Knowledge Editing Approaches} 
Recent advancements in knowledge editing have focused on updating pre-trained LLMs without re-training the entire model. These approaches can be broadly categorized as follows: 
\paragraph{Model-based Editing.} 
These approaches include parameter-centric techniques that directly modify model weights. For example, ROME (Rank-One Model Editing)~\cite{meng2022romeb} that treats the multi-layer perceptron (MLP) modules within transformers as key-value stores and applies rank-one updates to the MLP weights for precise factual updates. 
This method allows for targeted edits while maintaining the integrity of the model's internal knowledge. Similarly, MEMIT (Mass-Editing Memory in a Transformer)~\cite{meng2022memit} scales this process by updating multiple memories simultaneously and identifying critical layers for memory storage. Although this method excels in localized edits, it often struggles with maintaining consistency across interconnected facts.

\paragraph{External Memory Integration.} Approaches such as MeLLo (Memory-Based Language Model Editing)~\cite{zhong2024mquakeassessingknowledgeediting} decouple knowledge storage from model parameters by externalizing edited facts. By iteratively prompting LLMs to align outputs with these stored facts, MeLLo achieves robustness in multi-hop reasoning tasks. However, reliance on external memory introduces latency and scalability challenges.  

\paragraph{Modular Reasoning Frameworks.} Systems like PokeMQA (Programmable Knowledge Editing for Multi-hop QA)~\cite{gu2024pokemqa} decompose editing into subtasks such as question decomposition and conflict checking using specialized modules. This modular design improves interpretability but requires extensive task-specific engineering.  

\subsection{Benchmarks for Knowledge Editing}
Evaluation frameworks for knowledge editing fall into two paradigms:   
\paragraph{Logic-Agnostic Benchmarks.} These frameworks prioritize factual accuracy and edit locality over reasoning constraints. For example, Chen et al.\cite{chen2023knowedit} evaluate edits through reliability (post-edit correctness), generalization (paraphrase handling), and locality (unrelated fact preservation). Multi-hop benchmarks \cite{zhong2024mquakeassessingknowledgeediting} assess propagation of edits through knowledge chains, while programmable benchmarks~\cite{gu2024pokemqa} simulate dynamic scenarios. However, their synthetic nature limits real-world applicability.  

\paragraph{Logic-Aware Benchmarks.} These frameworks incorporate structured reasoning constraints. For example, the RULE-KE framework\cite{cheng2024leveraging} enforces consistency by leveraging logical rules to propagate edits across related facts. Ou et al. \cite{ou2025llmsacquirenewknowledge} use circuit-based analysis to localize logical inferences, while ontology-driven methods\cite{zhang2024comprehensive} verify hierarchical consistency via knowledge graphs. Wu et al. \cite{wu2024logicedit} extend this to temporal logic for dynamic knowledge. However, these benchmarks demand laborious rule formalization and struggle with scalability.  

\paragraph{Document-Based Benchmarks.} These frameworks test how well a model integrates knowledge edits in long-form text, with reasoning and generalization needs. For example, Wu et al. \cite{wu2025docterevaluatingdocumentbasedknowledge} introduces a new benchmark and evaluation framework for testing how well language models retain and apply edited knowledge in document-based settings, rather than just in isolated fact edits. However, its reasoning remain limited and lacks symbolic traceability. Moreover, the framework does not explicitly test for logical consistency or rule violations.
\begin{figure}[t!]
\centering
\includegraphics[width=\textwidth, trim={0.1cm 0 8.5cm 0},clip]{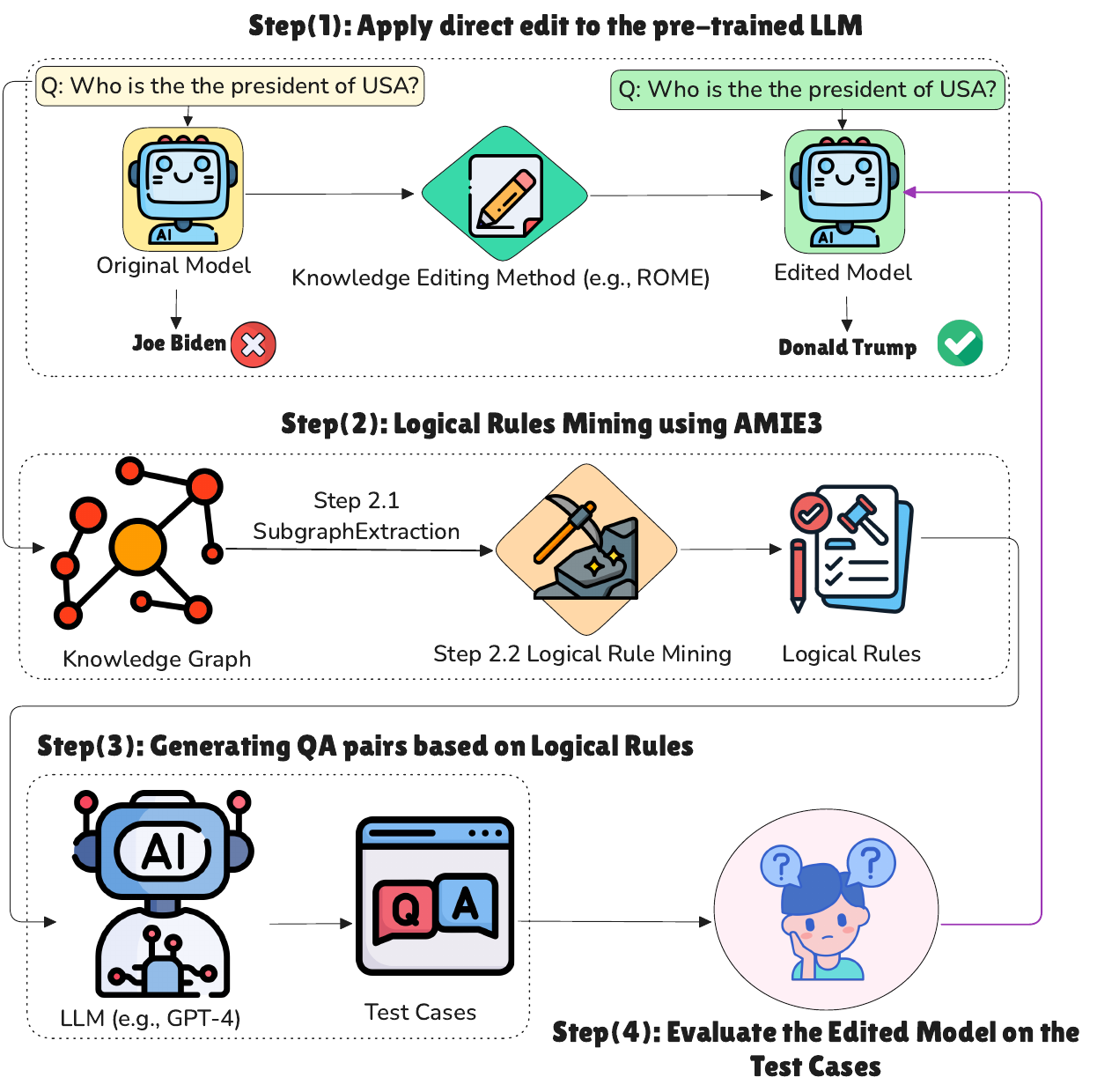}   
\caption{Our Benchmark Workflow for Evaluating Knowledge Editing Approaches.}
\label{fig:evaluation_workflow}
\end{figure}

\RestyleAlgo{ruled}
\SetKwComment{Comment}{/* }{ */}
\begin{algorithm*}[t!]
    \caption{Constructing $\mathcal{G}$}
    \label{algo:knowledge-graph-construction}
    \KwData{$D$ \tcp*{Benchmark dataset}}
    \KwResult{$\mathcal{G}$ \tcp*{The constructed KG (set of triples)}}
    $\mathcal{G} \gets \{\}$ \tcp*{Initialize set of triples}
    \ForEach{\texttt{question} in $D$}{
    $E \gets \text{spaCy}(\texttt{question})$ \tcp*{Extract entities from question}

    \ForEach{$e \in E$}{
        $e\_uri \gets find\_\text{DBpedia}\_uri(e)$\;
        \If{$e\_uri$ is valid}{
            $S_{triples} \gets query\_DBpedia\_\text{SPARQL}\_endpoint(e\_uri)$\;
            $\mathcal{G} \gets \mathcal{G} \cup S_{triples}$\;
        }
        \Else{
            skip triples related to $e\_uri$\;
        }
    }
    }

    \textbf{return} $\mathcal{G}$\;
\end{algorithm*}
\section{Methodology}
%In this section, we present the methodology employed to conduct all experiments.
%trim={0 4cm 0.5cm 0},clip
\subsection{Task Definition: Benchmarking Knowledge Editing in LLMs}
Knowledge editing updates assertional knowledge in language models (LLMs) while preserving their reasoning capabilities. We formalize this task through evaluation of direct edits and their correlated impacts. 
Let $\mathcal{G} \subseteq \mathcal{E} \times \mathcal{R} \times \mathcal{E}$ denote a knowledge graph with entities $\mathcal{E}$ and relations $\mathcal{R}$. An edit set $\mathcal{X} = \{(a_i, a'_i)\}_{i=1}^n$ contains assertion pairs where $a_i = (s_i, r_i, o_i)$ is the original triple and $a'_i = (s_i, r_i, o'_i)$ its update.We Given an LLM $L$ and editing method $\mathcal{M}$, we evaluate the edited model $L^* = \mathcal{M}(L, \mathcal{X})$ using two criteria:

%\an{Is $\mathbb{I}$ an indicator function? Notation rather unclear here.}
\begin{enumerate}
    \item \textbf{Edit Success}. The metric measures direct update effectiveness:
    \begin{equation}
    \label{eq:editsuccess}
        \text{EditSucc}(a_i, a'_i) = \mathbb{I}\left[P_{L^*}(o'_i \mid q(s_i, r_i)) > P_{L}(o_i \mid q(s_i, r_i))\right],
    \end{equation}
    where $q(s_i, r_i)$ queries $r_i$ about $s_i$, and $P_{L}(o_i \mid q)$ is $L$’s baseline probability for answer $o_i$.
    
    \item \textbf{Logical Consistency}. This ensures that edits propagate through logical dependencies. For each $(a_i, a'_i)$, let $R_i = \{\phi_j\}_{j=1}^k$ denote rules mined by AMIE3~\cite{10.1145/2488388.2488425}
    linking to correlated assertions $\{a^{(j)}_c = (s^{(j)}_c, r^{(j)}_c, o^{(j)}_c)\}$. We compute:
    \begin{equation}
    \label{eq:consistency}
        \text{Consistency}(a_i, a'_i) = \frac{1}{k}\sum_{j=1}^k \mathbb{I}\left[P_{L^*}(\hat{o}^{(j)}_c \mid q^{(j)}_c) > P_{L}(o^{(j)}_c \mid q^{(j)}_c)\right],
    \end{equation}
    where $\mathbb{I}$ is an indicator function that returns 1 if the condition is true, and 0 otherwise. $\hat{o}^{(j)}_c$ is the expected answer after applying rule $\phi_j$ to $a'_i$, and $q^{(j)}_c = q(s^{(j)}_c, r^{(j)}_c)$. For example, editing \textit{Valerie Hobson's citizenship} ($a_i$) generates queries about \textit{her husband's nationality} ($a^{(j)}_c$) via the rule $\phi_j: \text{spouse}(X,Y) \land \text{citizenship}(X,Z) \Rightarrow \text{nationality}(Y,Z)$.

\end{enumerate}

\subsection{Evaluation Workflow}
%\todo{HZ: rewrite this section to be aligned with the new figure, also add a descriptive caption for the figure}

\paragraph{Applying Direct Edit to the Pre-trained LLM}: As shown in \autoref{fig:evaluation_workflow}, we begin by applying the KE approach to the original model and save the edited model weights .

\paragraph{Mining Logical Rules}: We construct a knowledge graph and employ AMIE3 which serve as a mining tool to generate logical rules over a set of triples from the constructed knowledge graph.

\paragraph{Generating QA\_pairs based on the Rules}: We use the generated rules and prompt and LLM to generate QA\_pairs based on the provides rules which later serve as test cases to validate KE approaches. 

\RestyleAlgo{ruled}
\SetKwComment{Comment}{/* }{ */}
\begin{algorithm*}[t!]
\caption{QA\_pairs Generation.}
\label{alg:qa_pairs_generation}
\KwData{$D$, $\mathcal{G}$, $LLM$, $\Phi$ \tcp*{Dataset, knowledge graph, QA generation model, rules}}
\KwResult{$Q$ \tcp*{Test case questions and answers}}

$Q \gets \{\}$\tcp*{Initialize an empty dictionary}

\ForEach{\texttt{question} $\in D$}{
    $\texttt{triples} \gets \mathcal{G}[\texttt{question}]$ \tcp*{Get triples related to \texttt{question}}
    $\texttt{relevant\_relations} \gets \{r | (e_1, r, e_2)\in \texttt{triples}\}$\;
    %$relation\_rules \gets extract\_relation\_in\_rules()$ \tcp*{Assume this function is predefined}
    $\texttt{relevant\_rules} \gets \{\texttt{rule}\in \Phi\ | \text{ a } \texttt{relevant\_relation} \text{ appears in \texttt{rule}}\}$% \tcp*{Relevant rules}
    
    $\texttt{prompt} \gets get\_prompt(\texttt{relevant\_rules}, \texttt{triples})$\;

    \texttt{qa\_pair} $\gets$ $LLM$(\texttt{prompt})

    $Q$ $\gets$ $Q \cup \{\texttt{qa\_pair}\}$
}

\Return{$Q$}\;
\end{algorithm*}

\paragraph{Evaluating KE approaches on the Test Cases}: In this step, we consider the edited model and the generated test cases to assess KE techniques. Further details will be explained in the experiments section.

\section{Experiments}
We conducted our experiments to answer the following research questions:
\begin{itemize}
    \item \textbf{RQ}$_1$: \textit{How do LLMs respond logically to questions?}
    \item \textbf{RQ}$_2$: \textit{Are current knowledge editing techniques effective at modifying the logical reasoning capabilities of LLMs?} 
\end{itemize}

To address these questions, we benchmark existing KE techniques using logical rules generated from our knowledge graph, as follows:

\textit{Dataset Construction}. Given a dataset $D$ containing a collection of items, we extracted a set of entities $E$ from each question item and then identified their corresponding DBpedia URIs, which were used to construct the knowledge graph. 

\textit{Applying KE Techniques to LLMs.}  Since our goal is to evaluate whether a KE technique is effective at modifying knowledge within language models, we begin by applying each KE technique to the original dataset to edit the LLM’s knowledge. For each query in the dataset, we apply the corresponding KE technique and save the updated model weights along with the associated tokenizer. These components together represent the edited language model, denoted as $L^*$.

\textit{Knowledge Graph Construction}: As part of our benchmarking approach, we construct a knowledge graph $\mathcal{G}$ which is used to generate logical rules $\phi\in \Phi$ relevant for a given benchmark dataset. We summarize our knowledge graph construction procedure in Algorithm~\ref{algo:knowledge-graph-construction}. Starting from a given KE benchmark dataset $D$, we extract entities from questions using spaCy~\cite{vasiliev2020natural} and then use SPARQL~\cite{perez2009semantics} to retrieve triples involving those entities. Questions for which no entities could be identified were removed from our experiments. To build the knowledge graph, we employed \textit{SPARQLWrapper}\footnote{\url{https://sparqlwrapper.readthedocs.io/en/latest/main.html}}, a simple Python wrapper around a SPARQL service commonly used to remotely execute queries. Using this tool, we retrieved a set of triples $T$ which involve a given DBpedia URI. 

\textit{Rules Generation}: Once a knowledge graph is constructed using Algorithm~\ref{algo:knowledge-graph-construction}, we employ AMIE3~\footnote{\url{https://github.com/dig-team/amie}} to mine logical rules over the provided triples which were later used to assess the entailment reasoning capabilities of LLMs in responding to prompts. This process yields rules, which we use to generate new test cases for edited models, see below. 

\textit{Using LLMs to Generate Question-answer Pairs from Rules}: Here, we prompt a question generation model to generate our QA\_pairs based on the provided rules, as illustrated by Algorithm~\ref{alg:qa_pairs_generation}. For each original question in the dataset, we select rules that share the same predicate as the triples related to the question.\footnote{Here, triples for a question are obtained by using Algorithm~\ref{algo:knowledge-graph-construction}. The notation $\mathcal{G}[\texttt{question}]$ is used to refer to those triples.} Given a rule and a corresponding set of assertions, the task of the language model is to generate a multi-hop question whose answer is a single entity present in the assertions. Each QA\_pair includes the generated question, its answer, and the associated confidence score, as shown in Table~\ref{tab:qa_pairs}. The question should require reasoning that is grounded in the provided rule and the relationships among the assertions. For example, consider the rule: \texttt{bornIn(Person, Place) $\land$ livesIn(Person, Place) $\implies$ hasNationality(Person, Place)} and the assertion: ``John was born in German and lived in Frankfort since then''. A possible multi-hop question could be:
\textbf{Question}: ``What is the nationality of John?''. Answering this question requires multiple reasoning steps, e.g., \emph{1. Born in Germany, 2. Lives in Frankfort, 3. Frankfort is located within Germany}. The \textbf{Answer} to the question is indeed ``Germany''. Sample entries from the question-answer generation process are shown in Table\ref{tab:qa_pairs}.

\begin{table*}[t!]
\small
\centering
\caption{Sample of generated QA\_pairs with corresponding rules, assertions, and QA confidence scores.}
\resizebox{\textwidth}{!}{\begin{tabular}{p{2cm}|p{3cm}p{3cm}p{3cm}|p{1cm}}
\toprule
\textbf{Question} & \textbf{Rules} & \textbf{Assertions} & \textbf{QA\_pairs} & \textbf{Score} \\
\hline
\vspace{1cm}
\multirow{3}{=}{Where did Carl Sagan work during his career?} 
& "?a thumbnail ?h \newline ?b thumbnail ?h \newline $\implies$ ?a primaryTopic ?b" 
& "The\_Cosmic\_..."\newline "author" "Carl\_Sagan" 
& \textbf{Q:} What is Carl Sagan's occupation?\newline \textbf{A:} author 
& 0.98 \\
\cmidrule{2-5}

& "?g influencedBy ?b \newline ?g influences ?a \newline $\implies$ ?a primaryTopic ?b" 
& "Carl\_Sagan" \newline "subject" "Category:Cornell\_..." 
& \textbf{Q:} What is the name of the college where Carl Sagan studied?\newline \textbf{A:} Cornell University 
& 0.67 \\
\cmidrule{2-5}

& "?g influencedBy ?a \newline ?g influencedBy ?b \newline $\implies$ ?a primaryTopic ?b" 
& "Contact\_(novel)" \newline "author" "Carl\_Sagan" 
& \textbf{Q:} What is Carl Sagan's occupation?\newline \textbf{A:} author 
& 0.99 \\

\bottomrule
\end{tabular}
}
\label{tab:qa_pairs}
\end{table*}
The score indicates the confidence score that the model has in the answer it extracted from the context. A higher score indicates that the model is more confident that the extracted span of text is the correct answer to the given question within the provided context, while a lower score suggests that the model is less certain about the extracted answer.

\textit{Validating Edited Models against Rules}. Validating edited models against logical rules requires both the saved language model weights and the generated QA\_pairs. To assess a KE technique, we use ROME framework~\footnote{\url{https://github.com/kmeng01/rome}} as our baseline implementation. For each prompt in the dataset,  We first run each KE using the dataset and save the edited model $L^*$. Consequently, we prompt $L^*$ to predict an answer given a question from our QA\_pairs. The edited models should predict only the answer in the output that will be used to compare with the answer from the QA\_pairs thus, allowing to compute the F1 score. We then compute the F1 and EM scores to compare the predicted answers with the reference answers from the QA\_pairs. At the end our evaluation results include the generated questions and answers, the predicted answers and, the scores (F1 and EM). A sample of the evaluation results is available from our GitHub repository.\footnote{\url{https://github.com/dice-group/Benchmarking-KE/tree/main}}.  A low F1 score may indicate that the edit disrupted the model’s logical coherence, resulting in failures on related or entailed knowledge. Since Exact-Match measures whether the predicted answer exactly matches the ground truth (after normalization), its value is either $1$ (correct) or $0$ (incorrect), providing a binary signal of whether the edit was applied correctly.% The overall evaluation workflow is illustrated in Figure~\ref{fig:evaluation_workflow}.\todo{HZ: why do you mention this sentence here?, it seems incomplete idea}   

\subsection{Datasets}
In our experiments, we employ two benchmark datasets for knowledge editing:

\begin{itemize}
    \item \textbf{MLaKE}~\cite{wei2025mlake}, a multilingual knowledge editing benchmark designed for LLMs. The benchmark comprises $5,360$ single-hop and $4,072$ multi-hop questions generated from Wikipedia fact chains. It aims to evaluate the adaptability of LLMs to edits in their knowledge base and their ability to generalize the edited knowledge across different languages. In our study, we use the English single-hop free-form QA subset.% (see \ref{sec:mlake}).

    \item \textbf{MQuAKE}~\cite{zhong2024mquakeassessingknowledgeediting}, a multi-hop question answering benchmark for knowledge editing. It is designed to assess whether edited models can correctly answer questions whose answers should change as an entailed consequence of updated facts. In our study, we use the MQuAKE-T subset, which focuses on temporal knowledge updates. This dataset contains 1,825 instances and is intended to evaluate knowledge editing methods on real-world changes.% (see \ref{sec:mquake}).%\todo{missing citation}.

\end{itemize}

In each dataset, we retain key elements such as the \textit{question}, \textit{prompt}, \textit{subject} (extracted from the prompt), \textit{original answer}, and \textit{edited answer}. We use the question to extract entities that serve as the basis for constructing the knowledge graph, which in turn allows us to generate logical rules for evaluating the effectiveness of KE. The final data format used to generated qa\_pairs using MLaKE and MQuAKE comprises $4090$ and $4805$, respectively.

\subsection{Evaluation Metrics}
Here, we introduce metrics that are used to evaluate KE techniques. These metrics leverage Equations~ \ref{eq:editsuccess} and \ref{eq:consistency} to define answer correctness for direct edits and correlated assertions. We denote the answer produced by an edited model $L^*$ for a query $q$ by $ans(L^*,q)$. The metrics ``Exact Match'' and ``F1 Score'' are defined as follows.
%= Generated answer from edited model $L^*$ for query $q$. 

\begin{itemize}
    \item \textbf{Exact Match (EM)}:
    The model’s generated answer matches the expected answer exactly. For a set of test cases $\mathcal{T}$:
    \begin{equation}
        \text{EM} = \frac{1}{|\mathcal{T}|} \sum_{(q, \hat{o}) \in \mathcal{T}} \mathbb{I}\left[\text{ans}(L^*, q) = \hat{o}\right].
    \end{equation}
    
    \item \textbf{F1 Score}: 
    Token-level overlap between generated and expected answers. 
    For each test case, we tokenize $ans(L^*,q)$ and $ans_\text{expected}$ into sets of words, $T_\text{pred}$ and $T_{\text{true}}$, respectively: 

    %$ans_{\text{expected}}$
    \begin{align}
        \text{Precision} &= \frac{|T_{\text{pred}} \cap T_{\text{true}}|}{|T_{\text{pred}}|}, \quad
        \text{Recall} = \frac{|T_{\text{pred}} \cap T_{\text{true}}|}{|T_{\text{true}}|}, \\
        \text{F1} &= \frac{2 \cdot \text{Precision} \cdot \text{Recall}}{\text{Precision} + \text{Recall}}.
    \end{align}
\end{itemize}
\subsection{Baseline Methods and Models}
To implement our methodology in this study, we evaluate only parameter-based methods as baselines including Rank-One Model Editing (ROME), the Constrained Fine-tuning (FT)~\cite{zhu2020modifyingmemoriestransformermodels} and Knowledge Neurons (KN)~\cite{dai-etal-2022-knowledge}. For each KE technique, we conduct evaluations on directly edited knowledge (without injecting new knowledge), and correlated knowledge (QA\_pairs representing new knowledge).

\subsection{Experimental Setup}
To assess performance across different KE techniques, we conduct experiments using a set of models configured based on the ROME repository. For each model, we adjusted the hyperparameters to obtain more accurate results. For example, when running ROME on GPT2-large, we increased the \textit{v\_num\_grad\_steps} to 50. This parameter controls how many gradient accumulation steps are performed before a single weight update is applied, effectively allowing the model to incorporate targeted factual knowledge more precisely. It also enables finer-grained weight updates during the editing process and simulates large batch sizes, which is particularly useful when memory is limited.

Regarding layer configurations in ROME, we use layers 8, 12 and 17 as the targeted editing layers for GPT2-medium\footnote{\url{https://huggingface.co/openai-community/gpt2-medium}}, GPT2-large\footnote{\url{https://huggingface.co/openai-community/gpt2-large}}, and GPT2-xl\footnote{\url{https://huggingface.co/openai-community/gpt2-xl}} respectively, and layer 0 for all models in FT. All experiments were performed using PyTorch $2.4.1$ on an A100 80GB GPU.

\subsection{Results and Discussion}
In this section, we present the evaluation results of KE on the QA\_pairs generated based on logical rules. We construct our knowledge graph using a sample of $1000$ prompts (questions) from the MLaKE and MQuAKE datasets. We removed questions for which we could not find entities in the knowledge graph, resulting in a final set of $831$ questions for MLaKE and $961$ for MQuAKE. We compute the F1 score and EM for each generated QA\_pair. 
\newcolumntype{C}[1]{>{\centering\arraybackslash}p{#1}}
\begin{table*}[t!]
\centering
\caption{Evaluation results for ROME on direct (first part) and correlated knowledge (second part). Underlined scores are exception cases.}
\label{tab:ROME results}
%\resizebox{0.9\textwidth}{!}{
   \begin{tabularx}{0.7\textwidth}{p{3cm}XXXXX@{}}
        \toprule
        & \multicolumn{2}{c}{\textbf{MLaKE}} & \multicolumn{2}{c}{\textbf{MQuAKE}} \\
        \cmidrule(lr){2-3} \cmidrule(lr){4-5}
        \textbf{Models} & \textbf{F1} & \textbf{EM} & \textbf{F1} & \textbf{EM} \\
        \midrule
        gpt2-medium & 16.36 & 7.57 & 4.21 & 0.0 \\
        gpt2-large & 10.23 & 2.68 & 2.13 & 0.0 \\
        gpt2-xl & 12.90 & 3.78 & 1.51 & 0.0 \\
        \midrule
        gpt2-medium & 8.67 & 3.56 & 1.91 & 0.0 \\
        gpt2-large & 6.08 & 2.61 & \underline{2.42} & 0.0 \\
        gpt2-xl & 7.17 & 3.10 & \underline{3.84} & \underline{1.04} \\
        \bottomrule
    \end{tabularx}
%}
\end{table*}

\begin{table}[t!]
\centering
    \caption{Evaluation results for FT on direct (first part) and correlated knowledge (second part)}
    \label{tab:FT results}
    \begin{tabularx}{0.7\textwidth}{p{4cm}XXXXX@{}}
        \toprule
        & \multicolumn{2}{c}{\textbf{MLaKE}} & \multicolumn{2}{c}{\textbf{MQuAKE}} \\
        \cmidrule(lr){2-3} \cmidrule(lr){4-5}
        \textbf{Models} & \textbf{F1} & \textbf{EM} & \textbf{F1} & \textbf{EM} \\
        \midrule
        gpt2-medium\_constr & 15.18 & 1.83 & 4.97 & 0.0 \\
        gpt2-large\_constr & 24.58 & 7.45 & 9.10 & 0.0 \\
        gpt2-xl\_constr & 17.15 & 2.32 & 4.25 & 0.0 \\
        \midrule
        gpt2-medium\_constr & 0.90 & 0.04 & 0.008 & 0.0 \\
        gpt2-large\_constr & 0.45 & 0.12 & 0.28 & 0.0 \\
        gpt2-xl\_constr & 0.63 & 0.14 & 0.0 & 0.0 \\
        \bottomrule
    \end{tabularx}
\end{table}

\begin{table*}[t!]
\centering
    \caption{Evaluation results for KN on directly edited (direct) and correlated knowledge (indirect)}
    \label{tab:KN results}
        \begin{tabularx}{0.7\textwidth}{p{3cm}XXXXX@{}}
            \toprule
            & \multicolumn{2}{c}{\textbf{MLaKE}} & \multicolumn{2}{c}{\textbf{MQuAKE}} \\
            \cmidrule(lr){2-3} \cmidrule(lr){4-5}
            {\textbf{Models}} & F1 & EM & F1 & EM \\
            \midrule
            gpt2-xl\_direct & 1.34 & 0.0 & 4.67 & 3.44 \\
            \midrule
            gpt2-xl\_indirect & 14.26 & 6.54 & 18.53 & 10.23 \\
            \bottomrule
        \end{tabularx}
\end{table*}

\begin{figure*}[t!]
    \centering
        \includegraphics[width=0.99\textwidth,clip]{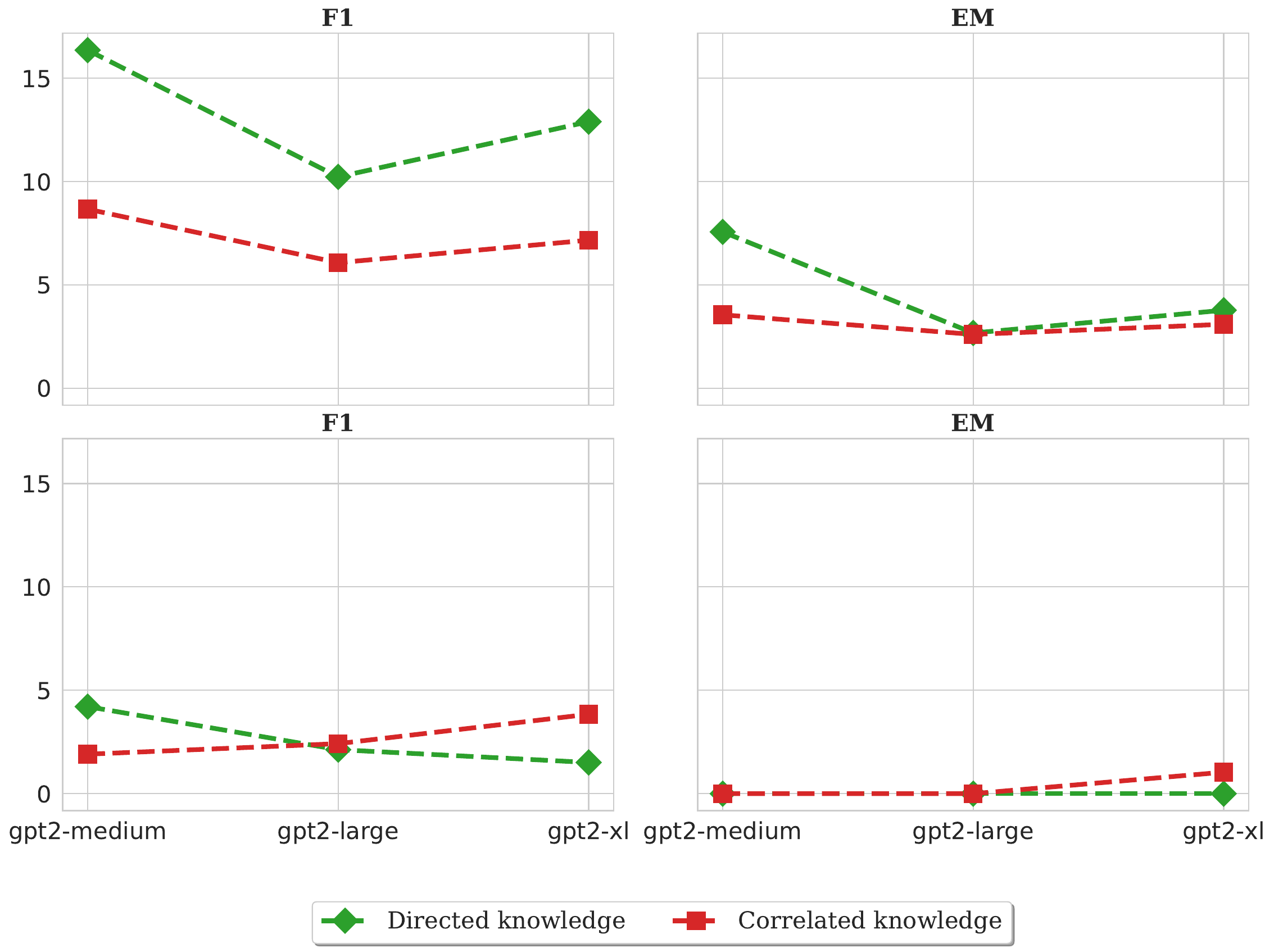}

    \caption{Evaluation scores (F1 on the left and EM on the right) of directed knowledge vs correlated knowledge of ROME using MLaKE(up) and MQuAKE(down)}

\label{fig:mlake_mquake_rome}
\end{figure*}

\paragraph{Results on ROME}: table \ref{tab:ROME results} presents the results on ROME when evaluating both directly edited knowledge and correlated knowledge using MLaKE and MQuAKE. We use gpt2-medium, gpt2-large, and gpt2-xl as the base models for editing. The results show that editing can disrupt the logical coherence of a model’s knowledge when evaluated against logical rules, except for gpt2-large and gpt2-xl using MQuAKE. Furthermore figure \ref{fig:mlake_mquake_rome} shows a performance decline, particularly with gpt2-large, indicating a failure of ROME to support logical reasoning. This suggests that either the neuron activations responsible for the model’s factual precision were not accurately identified, or the modified feed-forward weights used to update factual associations do not adequately account for entailments. As a result, the edited knowledge may fail to generalize properly and can lead to illogical conclusions in other contexts.

\begin{figure*}[t!]
    \centering
        \includegraphics[width=0.99\textwidth,clip]{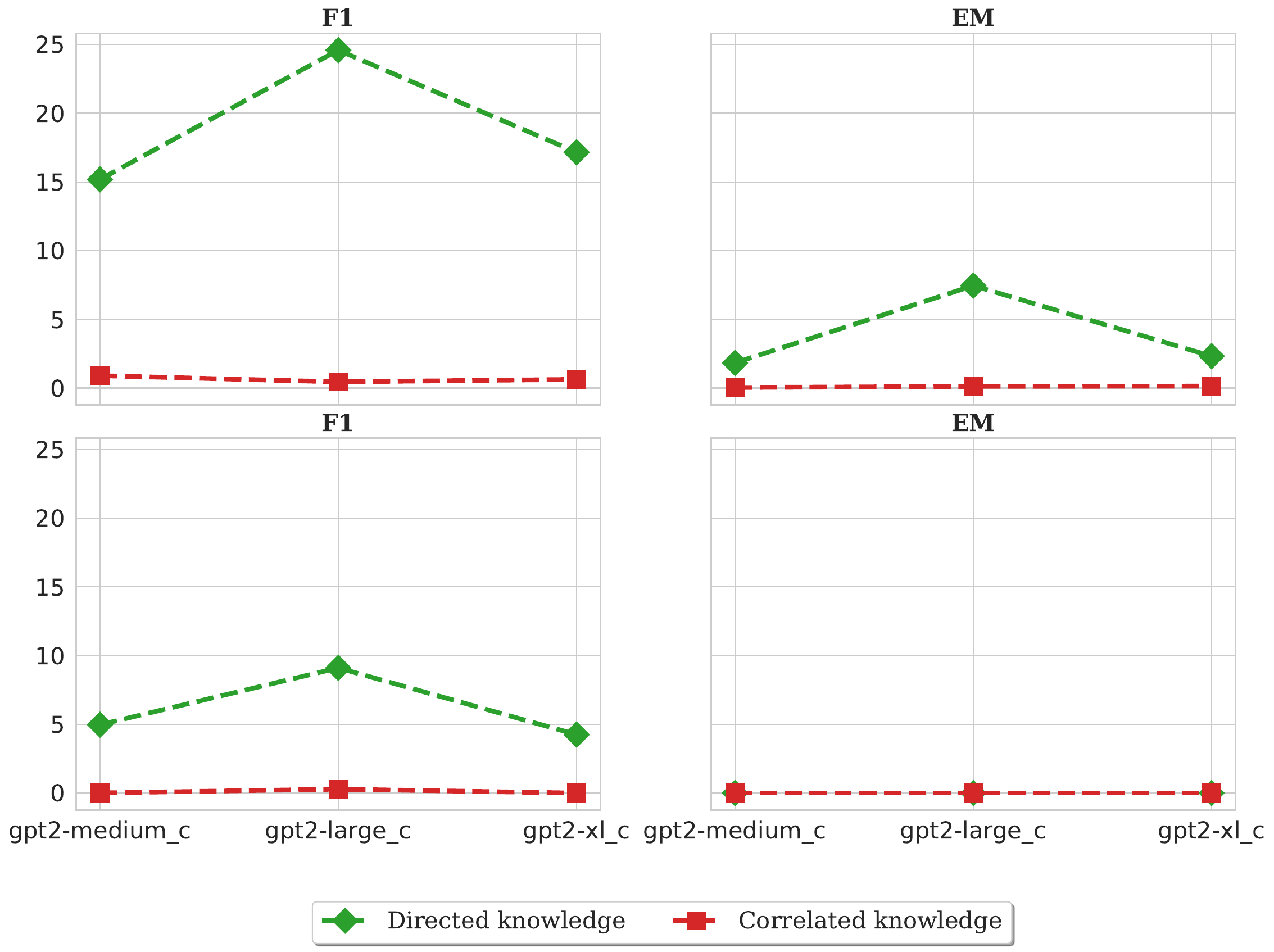}

    \caption{Evaluation scores (F1 on the left and EM on the right) of directed knowledge vs correlated knowledge of FT using MLaKE(up) and MQuAKE(down). The $\_c$ represents the constrains value added to the model. }
    \label{fig:ft_mlake_mquake}
\end{figure*}

\paragraph{Results on FT}: table \ref{tab:FT results} presents the results of FT when evaluating both directly edited knowledge and correlated knowledge using MLaKE and MQuAKE. As the base models for editing, we use gpt2-medium\_constr, gpt2-large\_constr, and gpt2-xl\_constr. We observe very low scores for all models when evaluating correlated knowledge, indicating that FT also fails to support entailment reasoning. The constrained FT approach involves fine-tuning the original model only on the modified facts while applying explicit constraints on the weights to minimize interference with unmodified facts. Additionally, a small positive constant is added as a constraint to the model’s performance. Moreover, figure \ref{fig:ft_mlake_mquake} shows a decline in accuracy on the modified assertions related to correlated knowledge, suggesting that the models undergo only minimal changes. This is likely due to the very small constraint values used during FT (0.0005, 0.001, and 0.002) for gpt2-xl\_constr, gpt2-large\_constr, and gpt2-medium\_constr, respectively). Although these constraints may significantly impact knowledge modification results, they do not necessarily indicate whether the models produce logically consistent conclusions in broader contexts.

\begin{figure*}[t!]
    \centering
        \includegraphics[width=0.99\textwidth,clip]{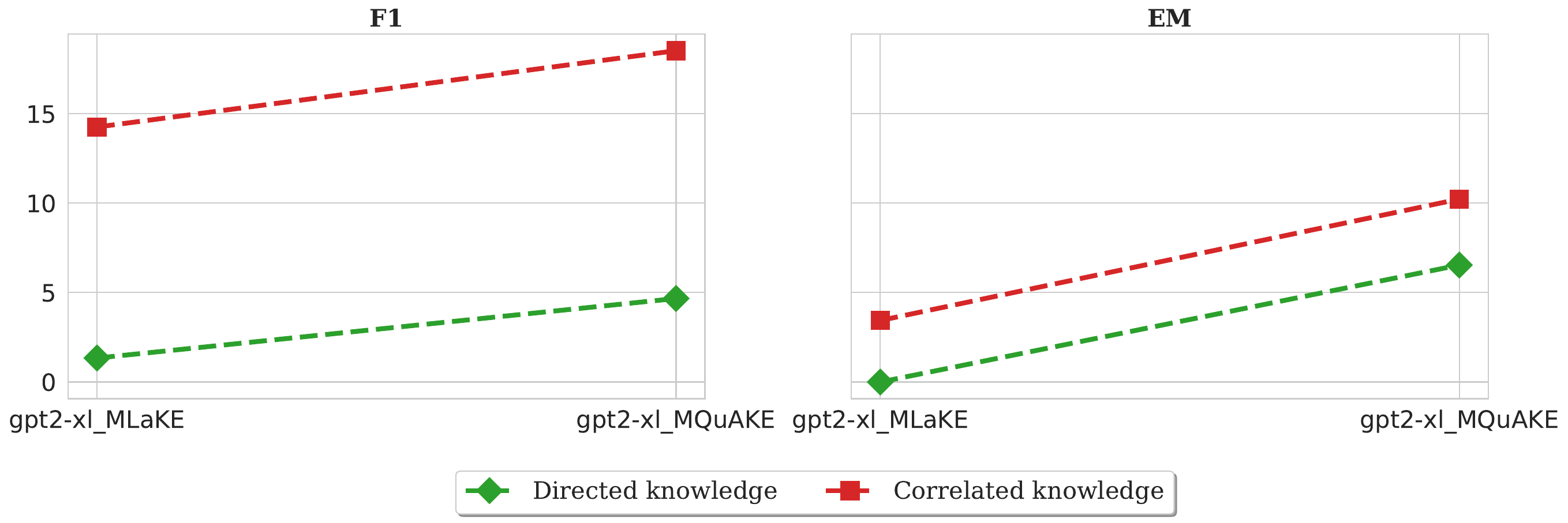}

    \caption{Evaluation scores (F1 on the left and EM on the right) of directed knowledge vs correlated knowledge of KN}
    \label{fig:kn_f1-em}
\end{figure*}

\paragraph{Results on KN}: table \ref{tab:KN results} presents the results of KN when evaluating both directly edited knowledge and correlated knowledge using MLaKE and MQuAKE. We use only gpt2-xl as the base model for editing. In contrast to other KE methods, KN performs relatively well when injecting correlated knowledge. KN involves identifying the neurons that represent the targeted fact and leveraging their activation to modify specific factual knowledge without full fine-tuning. As a result, the illustration in \ref{fig:kn_f1-em} demonstrates the effectiveness of this method in generalization, positively impacting the model's ability to perform reasoning tasks and handle broader knowledge domains.

\section{Conclusion and Future Work}

In this study, we propose a new benchmark for knowledge editing that includes multi-hop questions generated using logical rules to assess the effectiveness of existing KE methods. We evaluate editing performance using two benchmark datasets and find that methods such as ROME and FT do not adequately account for entailment and perform poorly when assessing the logical reasoning capabilities of LLMs. Interestingly, we observe more promising results with KN, which demonstrates better performance in generalizing correlated knowledge. Our findings reveal that although existing KE methods can accurately recall edited factual knowledge, they exhibit limitations when it comes to injecting correlated or broader domain knowledge. Furthermore, our benchmark helps enhance the evaluation of KE methods on multi-hop reasoning tasks. An important direction for future work is to extend our evaluation experiments on more recent LLMs such as LLaMA and Mistral, and very large knowledge graphs, such as Wikidata, by including other knowledge editing approaches (e.g., MeLLo), and explore how logical rules can further improve knowledge editing robustness and reasoning consistency.

\paragraph*{Resource Availability Statement:} Source code for our evaluation system is available from GitHub repository.\footnote{\url{https://github.com/dice-group/Benchmarking-KE/tree/main}}

\section*{Acknowledgment}
This work has been supported by the Ministry of Culture and Science of North Rhine-Westphalia (MKW NRW) within the project SAIL under the grant no NW21-059D and the German Federal Ministry of Research, Technology and Space (BMFTR) within the project KI-OWL under the grant no 01IS24057B. 

\bibliographystyle{splncs04}
\bibliography{main}

\end{document}